\documentclass[conference,letterpaper]{IEEEtran}
\IEEEoverridecommandlockouts
\usepackage{graphics,subfigure} 
\usepackage{epsfig} 
\usepackage{mathptmx} 
\usepackage{times} 
\usepackage{amsmath} 
\usepackage{amssymb}  
\usepackage{captcont}
\usepackage{graphicx}
\usepackage{multirow}
\usepackage{calc}
\usepackage{tikz}
\usepackage{algorithm}
\usepackage{algorithmic}
\usepackage{booktabs}
\usepackage{flushend}
\usepackage{slashbox}
\usepackage{cases}
\usepackage{array}
\usepackage{makecell}
\usepackage{bbm}
\usepackage{bm}
\usepackage{cite}
\usepackage[left=0.71in,top=0.94in,right=0.71in,bottom=1.18in]{geometry}

\makeatletter

\newcommand{\Rmnum}[1]{\expandafter\@slowromancap\romannumeral #1@}
\makeatother

\begin{document}
\title{\huge A Homotopy Coordinate Descent Optimization Method for $l_0$-Norm Regularized Least Square Problem
\footnoterule\thanks{$^{*}$This work is supported by National Natural Science Foundation of China (NSFC) under grant \#61873067.}
}

\author{\authorblockN{Zhenzhen Sun, Yuanlong Yu*}
\authorblockA{\textit{College of Mathematics and Computer Science}\\
\textit{Fuzhou University}\\
\textit{Fuzhou, Fujian, 350116, China}\\
\textit{yu.yuanlong@fzu.edu.cn}\\}%
}%

\maketitle
\begin{abstract}
This paper proposes a homotopy coordinate descent (HCD) method to solve the $l_0$-norm regularized least square ($l_0$-LS) problem for compressed sensing, which combine the homotopy technique with a a variant of coordinate descent method. Differs from the classical coordinate descent algorithms, HCD provides three strategies to speed up the convergence: warm start initialization, active set updating, and strong rule for active set initialization. The active set is pre-selected using a strong rule, then the coordinates of the active set are updated while those of inactive set are unchanged. The homotopy strategy provides a set of warm start initial solutions for a sequence of decreasing values of the regularization factor, which ensures all iterations along the homotopy solution path are sparse. Computational experiments on simulate signals and natural signals demonstrate effectiveness of the proposed algorithm, in accurately and efficiently reconstructing sparse solutions of the $l_0$-LS problem, whether the observation is noisy or not.
\\
\end{abstract}


\section{Introduction}
Sparse coding (SC) provides a class of algorithms for finding succinct representations of stimuli, which has been developed and applied in many fields over the last two decades, i.e., image denoising \cite{Chen15,Liu18}, feature selection \cite{Tibshirani:96,Pang19}, and pattern recognition \cite{Jiang15,Akhtar18}.

In general, sparse coding is based on the idea that, for an observed signal $\boldsymbol{x} \in R^d$ and an over-complete dictionary $\boldsymbol{D} \in R^{d \times K} (d \ll K)$, $\boldsymbol{x}$ can be reconstructed by a representation $\boldsymbol{\alpha} \in R^K$ using only a few atoms of the dictionary. More formally, the problem of finding the sparse representation $\boldsymbol{\alpha}$ is formulated as:
\begin{equation}\label{sparsecoding}
\begin{split}
\underset{\boldsymbol{\alpha}}{\min\text{:}} ||\boldsymbol{x}-\boldsymbol{D\alpha}||_2^2 +\lambda ||\boldsymbol{\alpha}||_0,
\end{split}
\end{equation}
where the $l_0$-norm is defined as the number of non-zero elements in a given vector, $\lambda$ is the regularization factor.

This problem has been proven to be a NP-hard problem, researchers have turned to approximately solve it instead. There are three common methods for approximations/relaxations of the problem: iterative greedy algorithms \cite{Mallat:93}, $l_1$-norm convex relaxation methods (which were called basis pursuit (BP)) \cite{Chen:01}, and $l_p$-norm ($0 < p <1$) relaxation methods \cite{Chartrand07,Xu12,Chen12,Qin13}. The most known greedy algorithms are orthogonal matching pursuit (OMP) \cite{Pati:93} and its  variations, i.e., StOMP \cite{Donoho:12}, MPL \cite{Tan:15a,Tan:15b}, RobOMP \cite{Loza:19}, etc.

BP method replace the $l_0$ norm with an $l_1$ norm to make a convex relaxation for the original problem, thus the objective function becomes:
\begin{equation}\label{BPDN}
\begin{split}
\underset{\boldsymbol{\alpha}}{\min\text{:}} ||\boldsymbol{x}-\boldsymbol{D\alpha}||_2^2 +\lambda ||\boldsymbol{\alpha}||_1,
\end{split}
\end{equation}
where the $l_1$-norm is defined as the sum of absolute values of all elements in a vector. This method has been proven to give the same solution to \eqref{sparsecoding} when the dictionary satisfies some conditions \cite{Donoho:03,Candes06}. Many research works have focused on efficiently solving problem~ \eqref{BPDN}, \cite{Allen:10} provides a comprehensive review of five representative methods, namely, \emph{Gradient Projection} (GP) \cite{Figueiredo:07,Kim:07}, \emph{Homotopy} \cite{Malioutov:05,Osborne:00}, \emph{Iterative Shrinkage-Thresholding} (IST) \cite{Combettes:05,Daubechies:04,E.T.Hale:07,Wright:09}, \emph{Proximal Gradient} (PG) \cite{beck:09,Lin:13}, and \emph{Augmented Lagrange Multiplier} (ALM) \cite{Yang:09}. Recently, a kind of pathwise coordinate optimization methods called PICASSO \cite{Zhao:17,Li:17,Ge:19} has been proposed to solve the $l_p$-LS ($0 < p \le 1$) problem, which has shown superior empirical performance than other state-of-the-art SC algorithms mentioned above.

Although satisfactory results can be achieved by the approximate/relax methods, from the sparsity perspective, $l_0$-norm is more desirable. In recent years, researchers try to solve problem \eqref{sparsecoding} directly, iterative hard thresholding (IHT) \cite{Blumensath:08,Blumensath:09,Lu:12} is the most popular method. The IHT methods have strong theoretical guarantees, and the extensive experimental results show that the IHT methods can be used to improve the results generated by other methods. Recently, Dong \emph{et al.} proposed two homotopy iterative hard-thresholding methods (HIHT and AHIHT) in \cite{Dong:18}, which combine the homotopy technique with IHT. The experimental results show that this two homotopy iterative hard-thresholding methods can improve the solution quality and speed up the convergence effectively.

However, IHT methods update all coordinates of $\boldsymbol{\alpha}$ in parallel thus they need to access all entries of the dictionary $\boldsymbol{D}$ in each iteration for computing a full gradient and a sophisticated line search step. Because of that, they are often not scalable and efficient in practice when $K$ is large. Inspired by PICASSO, this paper proposed a homotopy coordinate descent (HCD) method for solving problem \eqref{sparsecoding} directly, which combines the homotopy technique with a a variant of coordinate descent method to calculate and trace the solutions of the regularized problem along a continuous path. What's more, differing from the classical coordinate descent algorithms, HCD just update the coordinates of active set, which can reduce the computational time effectively, especially when the solution is very sparse. Experimental results show that the propose method is more efficient and effective than PICASSO and other homotopy methods.

The rest of this paper is organized as follows. Section \ref{Sec:Proposed} presents the proposed HCD method. In addition, convergence of this method is analyzed. Experimental results are presented in Section \ref{Sec:Exp} and the conclusions are made in Section \ref{Sec:Con}.
\section{The Proposed method}\label{Sec:Proposed}
\subsection{Problem Formulation}
For the sake of easy statement, problem \eqref{sparsecoding} is rewritten as:
\begin{equation}\label{ObjFun}
  \varPhi _{\lambda}\left( \boldsymbol{\alpha } \right) =\underset{\boldsymbol{\alpha }}{arg\min}\,\,\frac{1}{2}||\boldsymbol{x}-\boldsymbol{D\alpha ||}_{2}^{2}+\lambda ||\boldsymbol{\alpha} ||_0,
\end{equation}
since the square of $l_2$-norm and $l_0$-norm are all separable function, this problem can be optimized by classical coordinate descent optimization algorithm. Given $\boldsymbol{\alpha}^t$ at $t-th$ iteration, we select a coordinate $i$, and then take an exact coordinate minimization step
\begin{equation}\label{CoorObj}
  \varPhi _{\lambda}\left( \alpha _i \right) =\underset{\alpha _i}{arg\min}\frac{1}{2}||\boldsymbol{z}^t-\boldsymbol{d}_i\alpha _i||_{2}^{2}+\lambda ||\alpha _i||_0,
\end{equation}
where $\alpha_i$ is the $i-th$ element of $\boldsymbol{\alpha}$, $\boldsymbol{d}_i$ denotes the $i-th$ column of $\boldsymbol{D}$, and $\boldsymbol{z}^t=\boldsymbol{x}-\boldsymbol{D\alpha }^t+\boldsymbol{d}_i\alpha _{i}^{t}$ denotes the partial residual.

According to IHT, \eqref{CoorObj} admits a closed form solution computed by the hard thresholding operator \cite{Wright:09,Lu:12}:
\begin{equation}\label{HTO}
  \alpha _{i}^{t+1}=\varGamma _{\lambda ,i}\left( \boldsymbol{\alpha }^t \right) =\left\{ \begin{array}{c}
	s_{L_i}\left( \alpha _{i}^{t} \right), \ \ \  if\;||s_{L_i}\left( \alpha _{i}^{t} \right) ||_{2}^{2}>\frac{2\lambda}{L_i}\\
	\text{0,\;\,\,\,\,}if\;||s_{L_i}\left( \alpha _{i}^{t} \right) ||_{2}^{2}\leqslant \frac{2\lambda}{L_i}\;\;\;\;\;\\
\end{array} \right. ,
\end{equation}
where $s_{L_i}\left( \alpha _{i}^{t} \right) =\alpha _{i}^{t}-\frac{1}{L_i}\nabla f\left( \alpha _{i}^{t} \right)$, $\nabla f\left( \alpha _{i}^{t} \right)$ is the gradient of $f\left( \alpha _i \right) =\frac{1}{2}||\boldsymbol{z}^t-\boldsymbol{d}_i\alpha _i||_{2}^{2}$ which is Lipschitz continuous (denote its Lipschitz constant as $L_{f_i}$), constant $L_i > 0$ is an upper bound on the Lipschitz constant, i.e., $L_i \ge L_{f_i}$. We normalize each atom $\boldsymbol{d}_i$ of $\boldsymbol{D}$ such that $||\boldsymbol{d}_i||_2 = 1$, thus all $L_i$ can be set as $1$ (for convenience, we neglect the index $i$ and use $L$ to replace all $L_i$ in next).

\begin{figure*}[htb]
\centering
\includegraphics[scale=0.5]{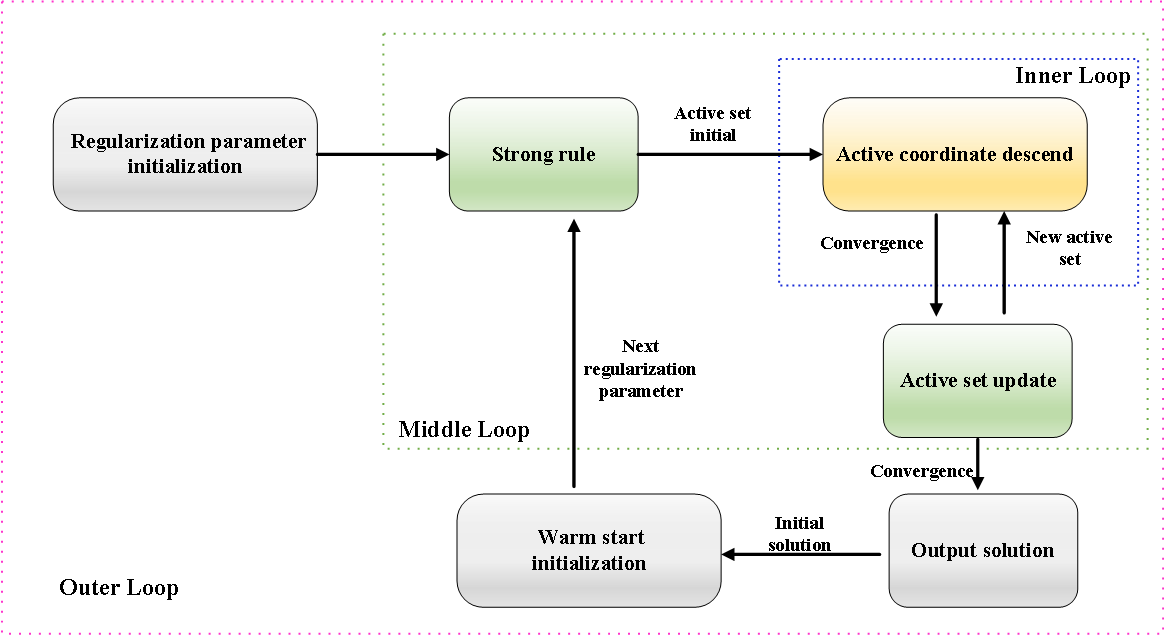}
\caption{Framework of HCD.}
\label{fig:frm}
\end{figure*}

It is time-consumption to update all coordinates, a homotopy coordinate descent optimization framework is proposed to the computational time, which integrates the warm start initialization, active set updating strategy, and strong rule for coordinate pre-selection into the classical coordinate optimization. These three strategies constitute three nested loops for the proposed algorithm, as illustrated in Fig.~\ref{fig:frm}. For simplicity, we first introduce its inner loop, then its middle loop, and at last its outer loop.
\subsection{Three Loops of HCD}
1. \emph{Inner Loop: Iterating over Coordinates within an Active Set}. The inner loop is denoted as active coordinate descent (ActCooDes) algorithm. The iteration index for the inner loop is $(t)$, where $t = 0, 1, 2, ...$. The ActCooDes algorithm solves \eqref{ObjFun} by conducting exact coordinate minimization iteratively, but it only updates a subset of all coordinates, which is called "active set". Accordingly, the complementary set to the active set is called "inactive set", whose coordinates do not change throughout all iterations of the inner loop. Since the active set usually contains few number of coordinates, ActCooDes algorithm is very scalable and efficient.

We denote the active and inactive sets as $A$ and $\bar{A}$ respectively. According to the initial solution of the inner loop $\boldsymbol{\alpha}^{(0)}$, active set and inactive set are selected based on the sparse pattern:
\begin{equation}\label{act}
  A=\left\{ j|\alpha _{j}^{\left( 0 \right)}\ne 0 \right\} \,\,and \,\, \bar{A}=\left\{ j|\alpha _{j}^{\left( 0 \right)}=0 \right\} .
\end{equation}

Then ActCooDes algorithm minimizes \eqref{ObjFun} with all coordinates of $\bar{A}$ keeping at zero values, the objective function becomes:
\begin{equation}\label{Objact}
  \underset{\boldsymbol{\alpha }\in R^d}{\min}\,\,\,\varPhi _{\lambda}\left( \boldsymbol{\alpha } \right), \,\, \text{subject to }\boldsymbol{\alpha }_{\bar{A}}=\mathbf{0}.
\end{equation}

Without loss of generality, we assume $\left| A \right|=s$, $A=\left\{ j_1...,j_s \right\} \subseteq \left\{1,...,K \right\}$, where $j_1\leqslant j_2\leqslant...\leqslant j_s$. Given a solution $\boldsymbol{\alpha}^{(t)}$ at $t-th$ iteration, we construct a sequence of auxiliary solutions $\left\{ \boldsymbol{w}^{\left( t+\text{1,}k \right)} \right\} _{k=0}^{s}$ to obtain $\boldsymbol{\alpha}^{(t+1)}$. Particularly, for $k = 0$, we have $\boldsymbol{w}^{\left( t+\text{1,}0 \right)}=\boldsymbol{\alpha }^{\left( t \right)}$; For $k = 1, ..., s$, we use \eqref{HTO} to update $w_{j_k}^{\left( t+\text{1,}k \right)}$ and make $\boldsymbol{w}_{\backslash j_k}^{\left( t+\text{1,}k \right)}=\boldsymbol{w}_{\backslash j_k}^{\left( t+\text{1,}k-1 \right)}$.

Then we set $\boldsymbol{\alpha}^{\left( t+1 \right)}=\boldsymbol{w}^{\left( t+\text{1,}s \right)}$ for the next iteration. The ActCooDes algorithm is terminated when
\begin{equation}\label{Term1}
  \lVert \boldsymbol{\alpha }^{\left( t+1 \right)}-\boldsymbol{\alpha }^{\left( t \right)} \rVert _2/||\boldsymbol{\alpha }^{\left( t \right)}||_2 < \tau \lambda ,
\end{equation}
where $\tau$ is a small convergence parameter (i.e., $10^{-5}$).

The outline of inner loop is presented in Algorithm~\ref{Alg:ActCooDes}
\begin{algorithm}[t]
\caption{$\boldsymbol{\hat{\alpha}}\gets ActCooDes\left(\boldsymbol{\alpha }^{\left( 0 \right)}, \lambda , \tau \right)$}
\label{Alg:ActCooDes}
\begin{algorithmic}
  \STATE \text{\textbf{(Input:)} $\boldsymbol{\alpha }^{\left( 0 \right)}, \lambda, \tau$}
  \STATE \text{initialize $t \gets 0, \ A=\left\{ j|\alpha _{j}^{\left( 0 \right)}\ne 0 \right\}$;} \\
          \text{\textbf{repeat}}\\
          \text{\ \ $\boldsymbol{w}^{\left( t+\text{1,}0 \right)}=\boldsymbol{\alpha }^{(t)}$;}\\
          \text{\ \ for $k=1:s$}\\
          \text{\ \ \ \ $w_{j_k}^{\left( t+\text{1,}k \right)}=\varGamma _{\lambda ,j_k}\left( \boldsymbol{w}^{\left( t+\text{1,}k-1 \right)} \right)$;}\\
          \text{\ \ \ \ $\boldsymbol{w}_{\backslash j_k}^{\left( t+\text{1,}k \right)}=\boldsymbol{w}_{\backslash j_k}^{\left( t+\text{1,}k-1 \right)}$}\\
          \text{\ \ end}\\
          \text{\ \ $\boldsymbol{\alpha }^{\left( t+1 \right)}\gets \boldsymbol{w}^{\left( t+\text{1,}s \right)}$;}\\
          \text{\ \ $t \gets t+1$;}\\
          \text{\textbf{until} $\lVert \boldsymbol{\alpha }^{\left( t+1 \right)}-\boldsymbol{\alpha }^{\left( t \right)} \rVert _2/||\boldsymbol{\alpha }^{\left( t \right)}||_2 < \tau \lambda $}\\
          \text{$\boldsymbol{\hat{\alpha}} \gets \boldsymbol{\alpha}^{(t)}$;}\\
\end{algorithmic}
\end{algorithm}

2. \emph{Middle Loop: Updating Active Sets Iteratively}. Since the inner loop can only converge to a local optimal solution of \eqref{Objact}, it is not necessarily the local optimal solution of \eqref{ObjFun}. Therefore, the inner loop needs to be combined with some active set updating scheme, which allows the active set to be changed. This leads to the middle loop of HCD.

The middle loop is denoted as  iterative active set updating (IteActUpd) algorithm, and the iteration index for it is $[m]$, where $m = 0, 1, 2, ...$. As shown in Algorithm \ref{Alg:IteActUpd}, the IteActUpd algorithm simultaneously updates the active set and decreases the objective value to ensure that the HCD algorithm converges to a local optimal solution of \eqref{ObjFun}.

(1) Strong Rule for Active Set Initialization: Tibshirani \emph{et al.} \cite{Tibshirani12} suggest a aggressive active set initialization procedure for PICASSO using a "strong rule", which often leads to superior computational performance in practice. Inspired by this, we also propose a "strong rule" for our HCD algorithm to initial active set. Suppose an initial solution $\boldsymbol{\alpha}^{[0]}$ is supplied to the middle loop, given an active set initialization parameter $\varphi \in \left( \text{0,}1 \right) $, the strong rule for HCD initializes $A_0$ and $\bar{A}_0$ as:
\begin{equation}\label{Iniact}
   \begin{array}{lll}
       A_0=\left\{ j|\alpha _{j}^{\left[ 0 \right]}=\text{0},|\nabla _jf\left( \boldsymbol{\alpha }^{\left[ 0 \right]} \right) |\geqslant \left( 1-\varphi \right) \sqrt{\frac{2\lambda}{L}} \right\}
       \\
       \,\,\,\,\,\,\,\,\,\,\,\,\,\,\,\,\,\,\,\      \cup \left\{ j|\alpha _{j}^{\left[ 0 \right]}\ne 0 \right\} ,
   \end{array}
\end{equation}

\begin{equation}\label{Iniinact}
  \bar{A}_0=\left\{ j|\alpha _{j}^{\left[ 0 \right]}=\text{0},|\nabla _jf\left( \boldsymbol{\alpha }^{\left[ 0 \right]} \right) |<\left( 1-\varphi \right) \sqrt{\frac{2\lambda}{L}} \right\},
\end{equation}
where $\nabla _jf\left( \boldsymbol{\alpha }^{\left[ 0 \right]} \right)$ denotes the $j-th$ entry of $\nabla f\left( \boldsymbol{\alpha }^{\left[ 0 \right]} \right)$, and $\nabla f\left( \boldsymbol{\alpha }\right)$ is the gradient of $f(\boldsymbol{\alpha}) = \frac{1}{2}||\boldsymbol{x} - \boldsymbol{D\alpha}||^2_2$. Note that the initialization parameters $\varphi$ need to be a reasonably small value (i.e., 0.1). Otherwise, the "strong rule" will choose too many active coordinates and affect the sparsity of the solution.

(2) Active Set Updating Strategy: Suppose at the $m-th$ iteration $(m \ge 1)$, we obtain a solution $\boldsymbol{\alpha}^{[m]}$ with a pair of active and inactive sets defined as:
\begin{equation}\label{act&inact}
  A_m=\left\{ j|\alpha _{j}^{\left[ m \right]}\ne 0 \right\}\,\,\ and \,\,\   \bar{A}_m=\left\{ j|\alpha _{j}^{\left[ m \right]}=0 \right\}.
\end{equation}

Each iteration of IteActUpd algorithm consists of two stages. The first stage is to conduct the ActCooDes algorithm over the active set $A_m$ until convergence, and then return a solution $\boldsymbol{\alpha }^{\left[ m+0.5 \right]}$. Since the coordinate descent algorithm may produce zero values for some active set coordinates, we remove these coordinates from the active set and update the active and inactive sets as follows:
\begin{equation}\label{Newact&inact}
   \begin{array}{lll}
       A_{m+0.5}=\left\{ j|\alpha _{j}^{\left[ {m+0.5} \right]}\ne 0 \right\},
       \\
       \bar{A}_{m+0.5}=\left\{ j|\alpha _{j}^{\left[ {m+0.5} \right]}=0 \right\}.
   \end{array}
\end{equation}

The second stage checks which coordinates of $\bar{A}_{m+0.5}$ should be added into the active set. We propose a greedy selection rule for updating the active set. Particularly, let $\nabla _jf\left( \boldsymbol{\alpha }^{\left[ m+0.5 \right]} \right) $ denote the $j-th$ entry of $\nabla f\left( \boldsymbol{\alpha }^{\left[ m+0.5 \right]} \right)$, we select a coordinate by
\begin{equation}\label{selcoo}
  k_m=argmax _{k\in \bar{A}_{m+0.5}}|\nabla _kf\left( \boldsymbol{\alpha }^{\left[ m+0.5 \right]} \right) |.
\end{equation}

The IteActUpd algorithm is terminated if
\begin{equation}\label{Term2}
  |\nabla _{k_m}f\left( \boldsymbol{\alpha }^{\left[ m+0.5 \right]} \right) |\leqslant \left( 1-\delta \right) \sqrt{\frac{2\lambda}{L}},
\end{equation}
where $\delta$ is a small convergence parameter (e.g., $10^{-5}$). Otherwise, we take
\begin{equation}\label{updcoo}
  \alpha _{k_m}^{\left[ m+1 \right]}=\varGamma _{\lambda ,k_m}\left( \boldsymbol{\alpha }^{\left[ m+0.5 \right]} \right) \,\, and \,\ \boldsymbol{\alpha }_{\backslash k_m}^{\left[ m+1 \right]}=\boldsymbol{\alpha }_{\backslash k_m}^{\left[ m+0.5 \right]},
\end{equation}
and update the active and inactive set as:
\begin{equation}\label{updact}
  A_{m+1}=A_{m+0.5}\cup \left\{ k_m \right\} \,\,\ and \,\,\,\ \bar{A}_{m+1}=\bar{A}_{m+0.5}\backslash \left\{ k_m \right\}.
\end{equation}

The outline of middle loop is presented in Algorithm~\ref{Alg:IteActUpd}.
\begin{algorithm}[t]
\caption{$\boldsymbol{\hat{\alpha}}\gets IteActUpd\left( \boldsymbol{\alpha }^{\left[ 0 \right]},\lambda ,\delta ,\tau ,\varphi \right) $}
\label{Alg:IteActUpd}
\begin{algorithmic}
  \STATE \text{\textbf{(Input:)} $\boldsymbol{\alpha }^{\left[ 0 \right]}, \lambda, \tau, \delta, \varphi$}
  \STATE \text{initialize $A_0=\left\{ j|\alpha _{j}^{\left[ 0 \right]}=\text{0},|\nabla _jf\left( \boldsymbol{\alpha }^{\left[ 0 \right]} \right) |\geqslant \left( 1-\varphi \right) \sqrt{\frac{2\lambda}{L}} \right\}$} \\
          \text{\ \ \ \ \ \ \ \ \ \ \ \ \ \ $ \cup \left\{ j|\alpha _{j}^{\left[ 0 \right]}\ne 0 \right\}, \,\,\ m \gets 0$;}\\
          \text{\textbf{repeat}}\\
          \text{\ \ $\boldsymbol{\alpha }^{\left[ m+0.5 \right]}\gets ActCooDes\left(\boldsymbol{\alpha }^{\left[ m \right]}, \lambda, \tau \right)$;}\\
          \text{\ \ $A_{m+0.5}\gets \left\{ j|\alpha _{j}^{\left[ m+0.5 \right]}\ne 0 \right\} , \bar{A}_{m+0.5}\gets \left\{ j|\alpha _{j}^{\left[ m+0.5 \right]}=0 \right\}$;}\\
          \text{\ \ $k_m=arg\max _{k\in \bar{A}_{m+0.5}}|\nabla _kf\left( \boldsymbol{\alpha }^{\left[ m+0.5 \right]} \right) |$;}\\
          \text{\ \ $\alpha _{k_m}^{\left[ m+1 \right]}\gets \varGamma _{\lambda ,k_m}\left( \boldsymbol{\alpha }^{\left[ m+0.5 \right]} \right) , \boldsymbol{\alpha }_{\backslash k_m}^{\left[ m+1 \right]}=\boldsymbol{\alpha }_{\backslash k_m}^{\left[ m+0.5 \right]}$;}\\
          \text{\ \ $A_{m+1}\gets A_{m+0.5}\cup \left\{ k_m \right\} , \bar{A}_{m+1}\gets \bar{A}_{m+0.5}\backslash \left\{ k_m \right\}$;}\\
          \text{\ \ $m \gets m+1$;}\\
          \text{\textbf{until} $|\nabla _{k_m}f\left( \boldsymbol{\alpha }^{\left[ m+0.5 \right]} \right) |\leqslant \left( 1-\delta \right) \sqrt{\frac{2\lambda}{L}}$}\\
          \text{$\boldsymbol{\hat{\alpha}} \gets \boldsymbol{\alpha}^{[m]}$;}\\
\end{algorithmic}
\end{algorithm}

3. \emph{Outer Loop: Iterating over Regularization Parameter}. The outer loop of HCD is the homotopy strategy which provides a warm starting initialization (WarStaInt). In the outer loop, we first set a large initial value of the regularization parameter $\lambda$ and gradually decrease it with a common ratio $\eta \in \left( \text{0,}1 \right)$ until it reach to the target value $\lambda_{tgt}$. For every fixed value of the regularization parameter, the IteActUpd algorithm is used to search an approximate optimal solution of \eqref{ObjFun}, which is set as the initial solution of the next iteration. Usually, the next loop with warm starting will require fewer iterations than current loop \cite{Wright:09}. We set the initial regularization parameter as $\lambda_0 = \lVert \boldsymbol{D}^T\boldsymbol{x} \rVert _{\infty}$ as with most algorithms. An outline of the WarStaInt algorithm is described as Algorithm 3.
\begin{algorithm}[t]
\caption{$\boldsymbol{\hat{\alpha}}\gets WarStaInt\left( \boldsymbol{\alpha }^0,\lambda _{tgt},\eta \right)$}
\label{Alg:WarStaInt}
\begin{algorithmic}
  \STATE \text{\textbf{(Input:)} $\boldsymbol{\alpha }^{0}, \lambda_{tgt}, \eta, \tau, \delta, \varphi$}
  \STATE \text{initialize $\lambda _0=||\nabla f\left( \boldsymbol{\alpha }^0 \right) ||_{\infty}, \ n \gets 0$;} \\
          \text{\textbf{repeat}}\\
          \text{\ \ $\lambda _{n+1}=\eta \lambda _{n}$;}\\
          \text{\ \ $\boldsymbol{\alpha}^{n+1}\gets IteActUpd\left(\boldsymbol{\alpha}^{n}, \lambda_{n+1}, \delta ,\tau ,\varphi \right) $;}\\
          \text{\ \ $n \gets n+1$;}\\
          \text{\textbf{until} $\lambda_{n}\le \lambda_{tgt}$}\\
          \text{$\boldsymbol{\hat{\alpha}} \gets \boldsymbol{\alpha}^{n}$;}\\
\end{algorithmic}
\end{algorithm}
\subsection{Convergence Analysis}
For a fixed $\lambda_n$, suppose the middle loop iterates to the $m-th$ time and $|A_m| = s$. For $i\in A_m$, it has been proven that \cite{Lu:12,Dong:18}
\begin{equation}\label{inequ1}
   \varPhi_{\lambda _n}\left( \alpha _{i}^{\left( t+1 \right)} \right) \leqslant  \varPhi_{\lambda _n}\left( \alpha _{i}^{\left( t \right)} \right),
\end{equation}
while for $i\in \bar{A}_m$, $ \varPhi _{\lambda _n}\left( \alpha _{i}^{(t+1)} \right) = \varPhi _{\lambda _n}\left( \alpha _{i}^{(t)} \right)$.

Thus, for all $\alpha _{i}^{(t+1)} \ (i=1,2,...,K)$, we  have
\begin{equation}\label{inequ2}
   \begin{array}{lll}
      \varPhi _{\lambda _n}\left( \boldsymbol{\alpha }^{(t+1)} \right) =\sum_{i=1}^K{ \varPhi _{\lambda _n}\left( \alpha _{i}^{(t+1)} \right)} \ \ \ \ \ \ \ \ \ \ \ \ \ \ \
      \\
      \,\,\,\,\,\,\,\,\,\,\,\,\,\,\,\,\,\,\,\,\,\,\,\,\,\,\,\,\,\,\,\,\,\,\,\,\,\,\,\,\,\,\,\,\,\,\,\ \leqslant \sum_{i=1}^K{ \varPhi_{\lambda _n}\left( \alpha _{i}^{(t)} \right)}=\varPhi _{\lambda _n}\left( \boldsymbol{\alpha }^{(t)} \right).
   \end{array}
\end{equation}

It implies that
\begin{equation}\label{inequ3}
  \begin{array}{lll}
      \varPhi _{\lambda _n}\left( \boldsymbol{\alpha }^{\left[ m+0.5 \right]} \right) =\varPhi _{\lambda _n}\left( \boldsymbol{\alpha }^{\left[ m+\text{0.5,}T_{m+0.5} \right]} \right) \ \ \ \ \ \ \
      \\
      \,\,\,\,\,\,\,\,\,\,\,\,\,\,\,\,\,\,\,\,\,\   \leqslant \varPhi _{\lambda _n}\left( \boldsymbol{\alpha }^{\left[ m+\text{0.5,}0 \right]} \right) =\varPhi _{\lambda _n}\left( \boldsymbol{\alpha }^{\left[ m \right]} \right) ,
  \end{array}
\end{equation}
where $T_{m+0.5}$ is the number of iterations of ActCooDes in $m-th$ iteration of middle loop.

Similarly, we have
\begin{equation}\label{inequ4}
  \varPhi _{\lambda _n}\left( \boldsymbol{\alpha }^{\left[ m+1 \right]} \right) \leqslant \varPhi _{\lambda _n}\left( \boldsymbol{\alpha }^{\left[ m+0.5 \right]} \right) \leqslant \varPhi _{\lambda _n}\left( \boldsymbol{\alpha }^{\left[ m \right]} \right),
\end{equation}
this inequality implies that, for a fixed $\lambda_n$, $\varPhi _{\lambda _n}\left\{ \boldsymbol{\alpha }^{\left[ m \right]} \right\}$ is nonincreasing. Since $f(\boldsymbol{\alpha})$ is bounded below, it then follows that ${\varPhi}_{\lambda_n}\left\{ \boldsymbol{\alpha }^{\left[ m \right]} \right\}$ is bounded below. Hence, ${\varPhi}_{\lambda_n}\left\{ \boldsymbol{\alpha }^{\left[ m \right]} \right\}$ converges to a finite value as $m\rightarrow \infty $ and a local optimal solution $\boldsymbol{\alpha }_{\lambda_n}^*$ can be achieved.

Since the $\lambda$ is monotone decreased, and $\boldsymbol{\alpha }_{\lambda_n}^*$ is set as the initial solution for IteActUpd in $\lambda_{n+1}$, we obtain that:
\begin{equation}\label{inequ5}
  \varPhi _{\lambda _n}\left( \boldsymbol{\alpha }_{\lambda _n}^{*} \right) >\varPhi _{\lambda _{n+1}}\left( \boldsymbol{\alpha }_{\lambda _n}^{*} \right) =\varPhi _{\lambda _{n+1}}\left( \boldsymbol{\alpha }_{\lambda _{n+1}}^{0} \right) \geqslant \varPhi _{\lambda _{n+1}}\left( \boldsymbol{\alpha }_{\lambda _{n+1}}^{*} \right) ,
\end{equation}
it implies that the objective value is monotone decreasing and a local optimal solution can be achieved by the proposed algorithm.
\section{Experiment}\label{Sec:Exp}
In this section, we conduct computational experiments for testing the performances of our HCD method on
reconstructing sparse representation for observed signal. The proposed method is compared with PICASSO and three state-of-the-art homotopy algorithms, namely PGH \cite{Lin:13}, HIHT \cite{Dong:18} and AHIHT \cite{Dong:18}. All experiments are performed on a personal computer with an Intel $Core^{TM}$ i7-7700 CPU (3.60 GHz) and 32-GB memory, using a MATLAB toolbox.

The experiments are mainly divide into three parts: (1) We evaluate the effectiveness of our algorithm; (2) We evaluate the influence of parameters $\lambda_{tgt}$ and $\eta$ in the algorithm; (3) We compare the performance of our algorithm with the compared algorithms in generated signals and natural signals.

Suppose $\boldsymbol{\hat{\alpha}}$ is the obtained solution, $\boldsymbol{\alpha}^*$ is the unknown sparse representation, the validation metrics used in our experiments include:
\begin{enumerate}
  \item Reconstruction error: $\varepsilon =\lVert \boldsymbol{x}-\boldsymbol{D\hat{\alpha}} \rVert _2$;
  \item Objective gap: $obj\_gap\text{=}\varPhi \left( \boldsymbol{\hat{\alpha}} \right) -\varPhi ^*$, where $\varPhi ^*=\varPhi \left( \boldsymbol{\alpha }^* \right) $;
  \item Sparsity: $nnz=\lVert \boldsymbol{\hat{\alpha}} \rVert _0$;
  \item Reconstruction time: CPU time (in seconds).
\end{enumerate}
\subsection{Data Generation and Parameter Setting}
In the experiments, we use normal distribution and uniform distribution to generate simulate signals. For normal distribution, we firstly randomly generated the dictionary $\boldsymbol{D} \in R^{d\times K}$ with mean 0 and  standard deviation 1, the vector $\boldsymbol{\alpha}^* \in R^K$ was generated with the same distribution at $s$ randomly chosen coordinates ($||\boldsymbol{\alpha}^*||_0 = s$), the noise $\boldsymbol{z} \in R^d$ is a dense vector with independent random entries with mean 0 and  standard deviation $\sigma $. For uniform distribution, the entries of $\boldsymbol{D}$ are generated independently with the uniform distribution over the interval $[-1, +1]$, the noise is a dense vector with independent random entries with the uniform distribution over the interval $[-\sigma, \sigma]$, where $\sigma$ is the noise magnitude. Finally, the observed signal $\boldsymbol{x} \in R^d$ is generated as $\boldsymbol{x} = \boldsymbol{D\alpha^{*}} +\boldsymbol{z}$.

For natural signals, we randomly extracted $10$ image patches from Barbara and Lena images to generate natural signals. Barbara image is noise-free and Lena image has random gaussian noise with $\sigma =10$. The patch size of Barbara and Lena are $8 \times 8$ and $16 \times 16$, respectively (it means the dimension $d$ of observed signal are 64 and 256, respectively). The dictionary is randomly generated with normal distribution and be normalized, the number of atoms are set as $K = 256$ and $K = 1024$, respectively. The average results of $10$ image patches are recorded for comparison.

In the experiments, unless otherwise stated, all parameters are set as follows. The initial value $\lambda_0$ of all algorithms is set as $\lVert \boldsymbol{D}^T\boldsymbol{x} \rVert _{\infty}$, the initial solution is set as $\boldsymbol{\alpha}^0 = \boldsymbol{0}$. Other parameters are set as: $\tau =10^{-6}$, $\delta =10^{-3}$, $\varphi =0.05$, $\eta=0.5$.
\subsection{The Effectiveness of HCD}
\begin{figure*}[htp]
\centering
\subfigure[]{
\label{Fig:scatter}\includegraphics[width=7.5cm,height=6cm]{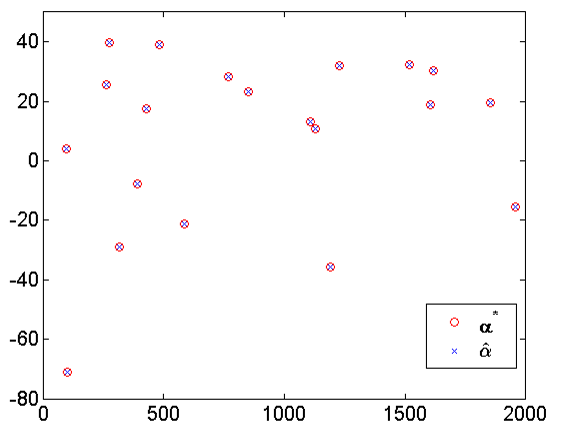}}
\subfigure[]{
\label{Fig:convergence}\includegraphics[width=7.5cm,height=6cm]{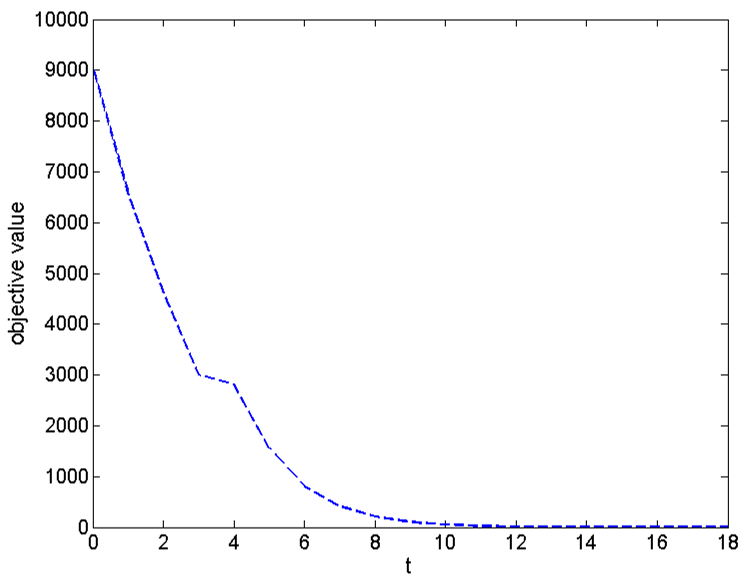}}
\caption{Reconstruction Results (Case One). \subref{Fig:scatter} Scatter diagram of original sparse signal and reconstructed signal. \subref{Fig:convergence} Convergence curve.}
\label{fig:RR1}
\end{figure*}

\begin{figure*}[htp]
\centering
\subfigure[]{
\label{Fig:scatter1}\includegraphics[width=7.5cm,height=6cm]{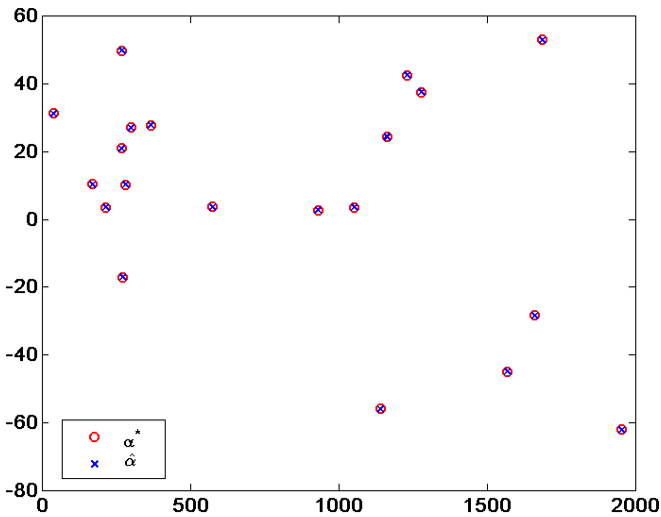}}
\subfigure[]{
\label{Fig:convergence1}\includegraphics[width=7.5cm,height=6cm]{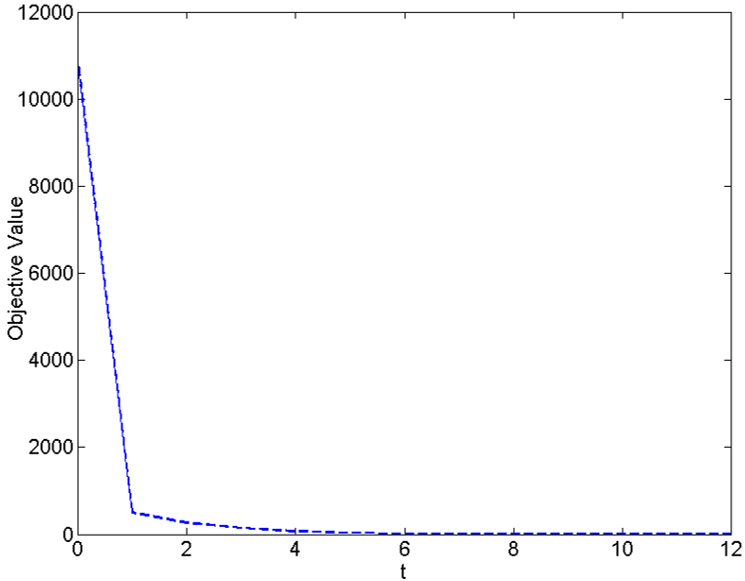}}
\caption{Reconstruction Results (Case Two). \subref{Fig:scatter} Scatter diagram of original sparse signal and reconstructed signal. \subref{Fig:convergence} Convergence curve.}
\label{fig:RR2}
\end{figure*}

In this part, we generated normal distributed noise-free signal with $(d=300, K=2000, s=20)$ to verify the proposed method, the regularization parameter $\lambda_{tgt}$ is set as $0.01$, 20 repeated trials are carried out and average results are recorded. Fig.~\ref{fig:RR1} and Fig.~\ref{fig:RR2} show two cases of the scatter diagram of reconstructed signal and convergence curve, and Tab.~\ref{performance results} presents the average results on the validation metrics. As it can be seen from Fig.~\ref{Fig:scatter} and Fig.~\ref{Fig:scatter1} that the solution $\boldsymbol{\hat{\alpha}}$ obtained by our algorithm coincides completely with the original sparse signal $\boldsymbol{\alpha}^*$, which indicates that our algorithm can achieve the optimal solution and has strong reconstruction performance, the results presented in Tab.~\ref{performance results} also indicate the effectiveness of our algorithm. From Fig.~\ref{Fig:convergence} and Fig.~\ref{Fig:convergence1} we can see that, only a dozen iterations can our algorithm convergent to the terminate condition, and from Tab.~\ref{performance results} we can see the average time spent of our algorithm to reconstruct the sparse signal is only $0.6074$ seconds. This two results indicate that the convergence speed of our algorithm is very fast which is applicable in practice.

\begin{table}[htp]
\caption{Other Performance Indicators Results}
\centering
\begin{tabular}{cc}
\hline
{Metric}&{Value}\\\hline
{$\varepsilon$}&{$5.1611e-9$}\\
{$obj\_gap$}&{$9.8879e-17$}\\
{$nnz$}&{20}\\
{CPU times}&{0.6074}\\\hline
\end{tabular}
\label{performance results}
\end{table}
\subsection{Parameter Sensitivity}
\begin{figure*}[htp]
\centering
\subfigure[]{
\label{Fig:lambdaofobj-gap}\includegraphics[width=7.5cm,height=6cm]{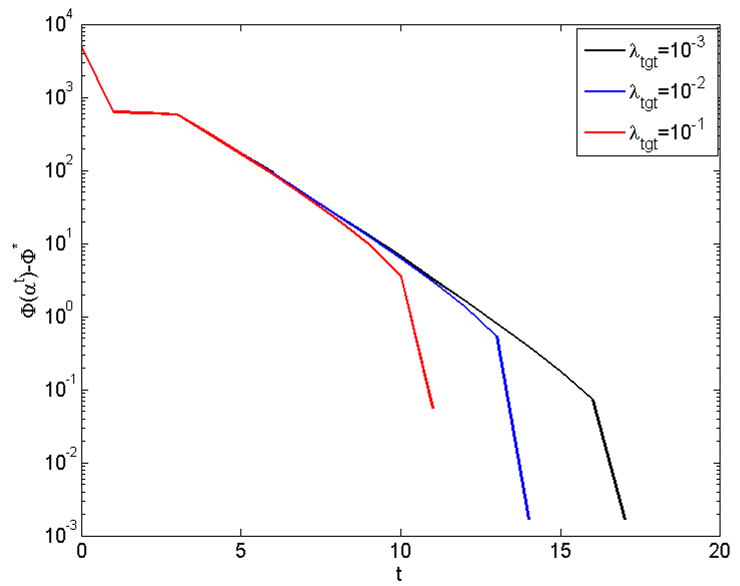}}
\subfigure[]{
\label{Fig:lambdaofsparsity}\includegraphics[width=7.5cm,height=6cm]{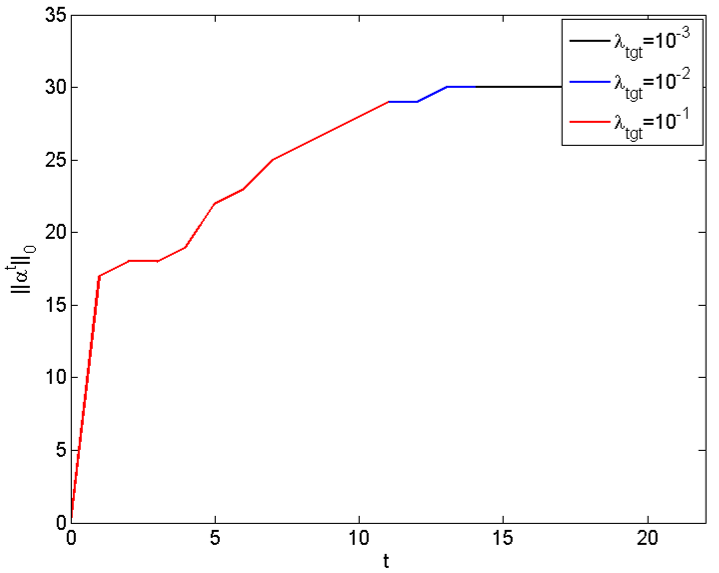}}
\caption{Performance of the HCD method by varying $\lambda_{tgt}$. \subref{Fig:lambdaofobj-gap}: $obj\_gap$. \subref{Fig:lambdaofsparsity}:Sparsity along solution path.}
\label{fig:influence of lambda}
\end{figure*}

In this part, we investigate the sensitivity of parameters $\lambda_{tgt}$ and $\eta$ in HCD. First, the normal distributed noisy signal with $(d=400, K=2000, s=30, \sigma=0.01)$ is generated to evaluate the influence of regularization parameter $\lambda_{tgt}$, the results with different values of $\lambda_{tgt}$ are shown in Fig.~\ref{fig:influence of lambda}. From this figure it can be seen that when $\lambda_{tgt} = 10^{-3}$ or $10^{-2}$, HCD can obtain the same solution. While when $\lambda_{tgt} = 10^{-1}$, the solution obtained is a litter more sparse than original sparse signal and results in a large objective gap, but is still acceptable. Therefore, within a certain range of $\lambda_{tgt}$, it only has a great influence on the number of iterations which will not influence the final result. In particular, it can be seen from Fig.\ref{Fig:lambdaofsparsity} that, despite $\lambda_{tgt}$ is changed, HCD traced the same solution path in previous iterations. The experimental results show that the proposed algorithm is robust to the regularization parameter, and it is unnecessary to spend too much time on parameter tuning.

\begin{figure*}[htp]
\centering
\subfigure[]{
\label{Fig:etaofobj-gap}\includegraphics[width=7.5cm,height=6cm]{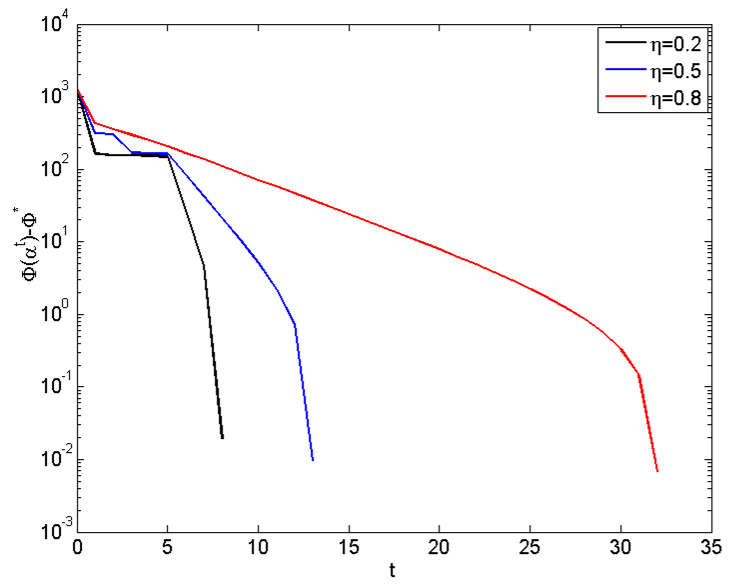}}
\subfigure[]{
\label{Fig:etaofsparsity}\includegraphics[width=7.5cm,height=6cm]{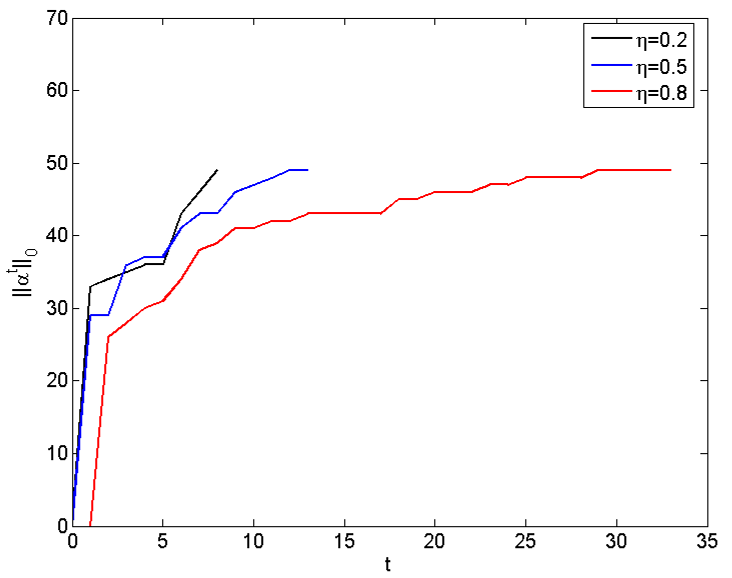}}
\caption{Performance of the HCD method by varying $\eta$. \subref{Fig:etaofobj-gap}: $obj\_gap$. \subref{Fig:etaofsparsity}:Sparsity along solution path.}
\label{fig:influence of eta}
\end{figure*}

In order to evaluate the influence of $\eta$ in HCD, we generated uniform distributed noisy signal with $(d=500, K=2000, s=50, \sigma=0.01)$ to do the experiment, the results with different values of $\eta$ are shown in Fig.~\ref{fig:influence of eta}. It can be seen from this figure that, though different values of $\eta$ can get similar results, the number of iterations of the algorithm gradually increases as $\eta$ increases. However, if $\eta$ is too small (i.e., 0.2), the gap between two adjacent values of $\lambda$ is too large, which will make HCD take more time to search for the solution (the CPU time of HCD with respect to these three values of $\eta$ are $2.44s$, $1.47s$ and $1.79s$, respectively). Therefore, for the following experiments, $\eta$ is set as $0.5$.
\subsection{Comparison Results}
\begin{figure*}[htp]
\centering
\subfigure[]{
\label{Fig:obj-gap32}\includegraphics[width=7.5cm,height=6cm]{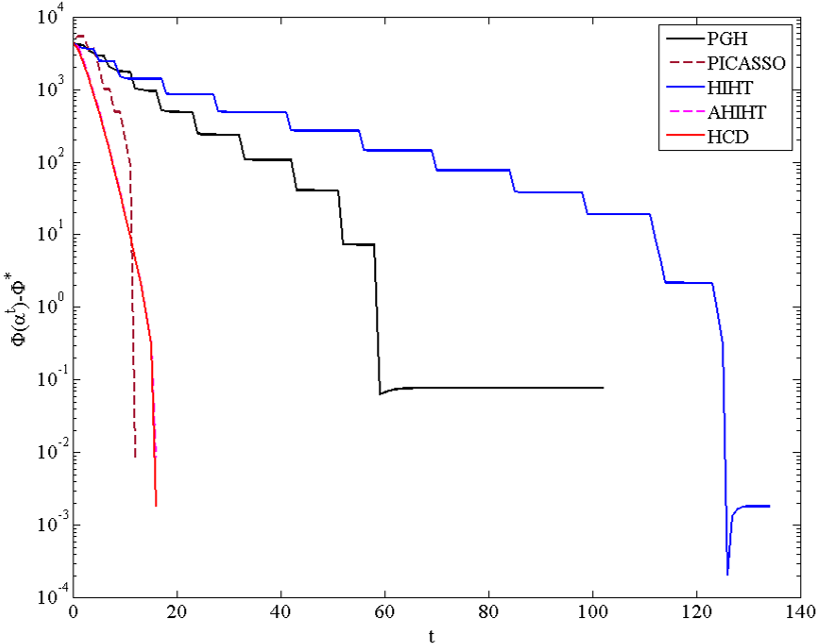}}
\subfigure[]{
\label{Fig:err32}\includegraphics[width=7.5cm,height=6cm]{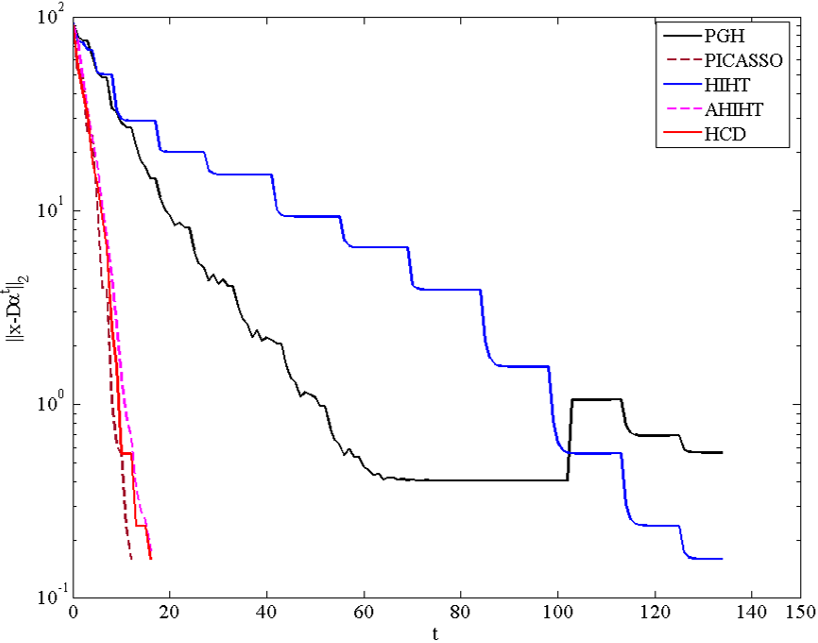}}
\subfigure[]{
\label{Fig:sparsity32}\includegraphics[width=7.5cm,height=6cm]{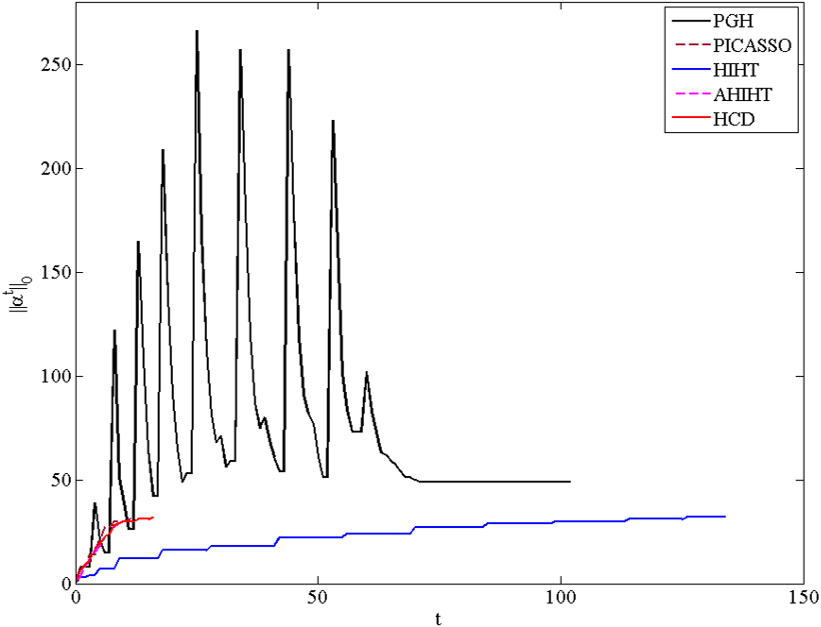}}
\subfigure[]{
\label{Fig:iteration32}\includegraphics[width=7.5cm,height=6cm]{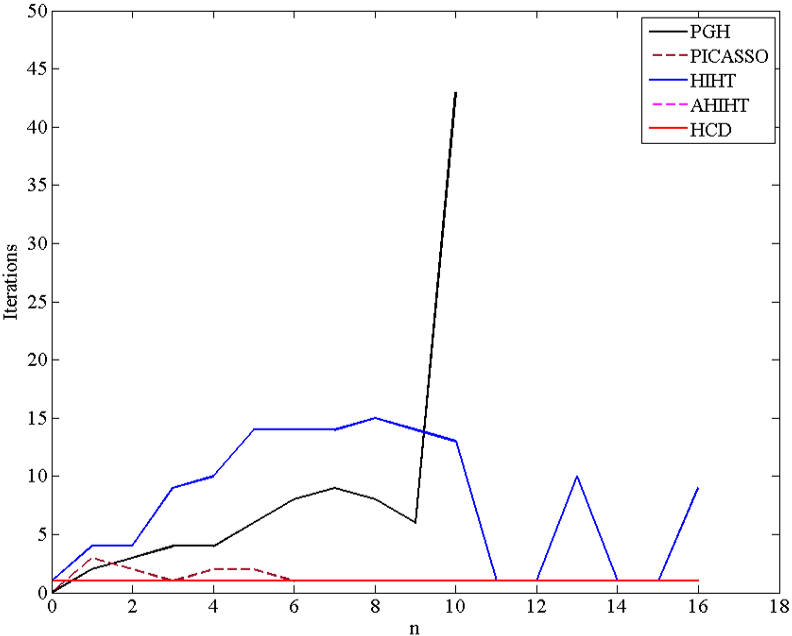}}
\caption{Performance of the compared methods on normal distributed noisy signal. \subref{Fig:obj-gap32}: $obj\_gap$. \subref{Fig:err32}: Reconstruction error. \subref{Fig:sparsity32}: Sparsity along solution path. \subref{Fig:iteration32}: Number of iterations of each $\lambda_n$.}
\label{fig:Performance of normal distributed noisy signal}
\end{figure*}

\begin{figure*}[htp]
\centering
\subfigure[]{
\label{Fig:obj-gap100}\includegraphics[width=7.5cm,height=6cm]{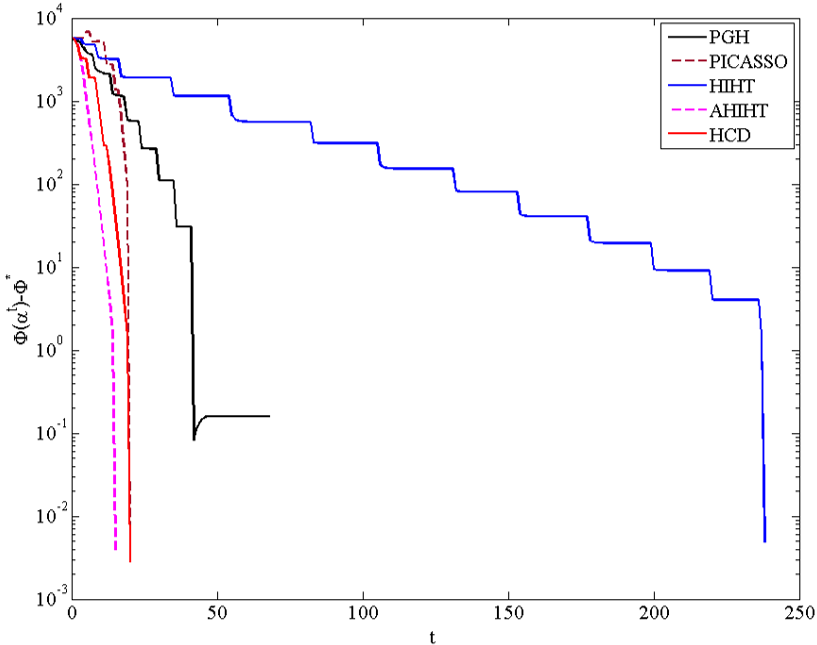}}
\subfigure[]{
\label{Fig:err100}\includegraphics[width=7.5cm,height=6cm]{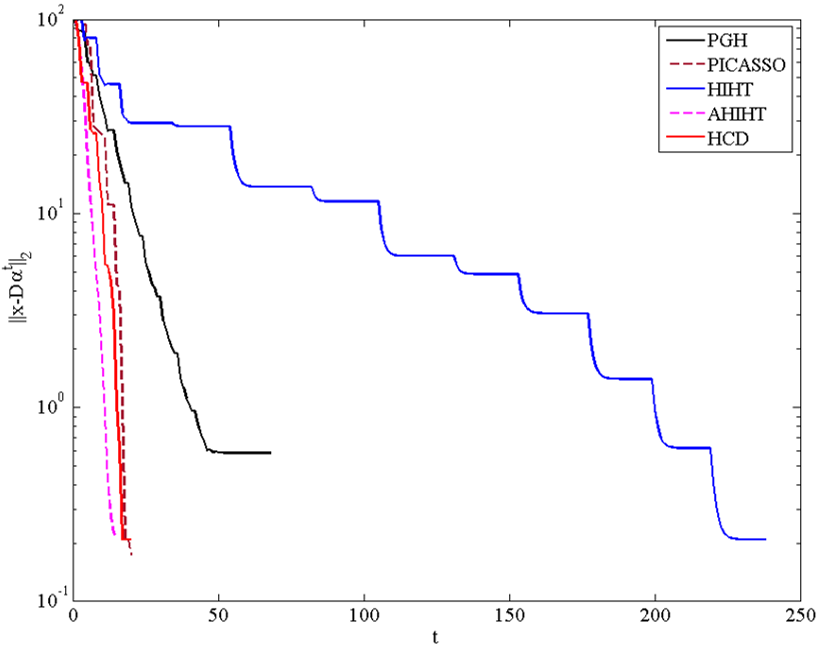}}
\subfigure[]{
\label{Fig:sparsity100}\includegraphics[width=7.5cm,height=6cm]{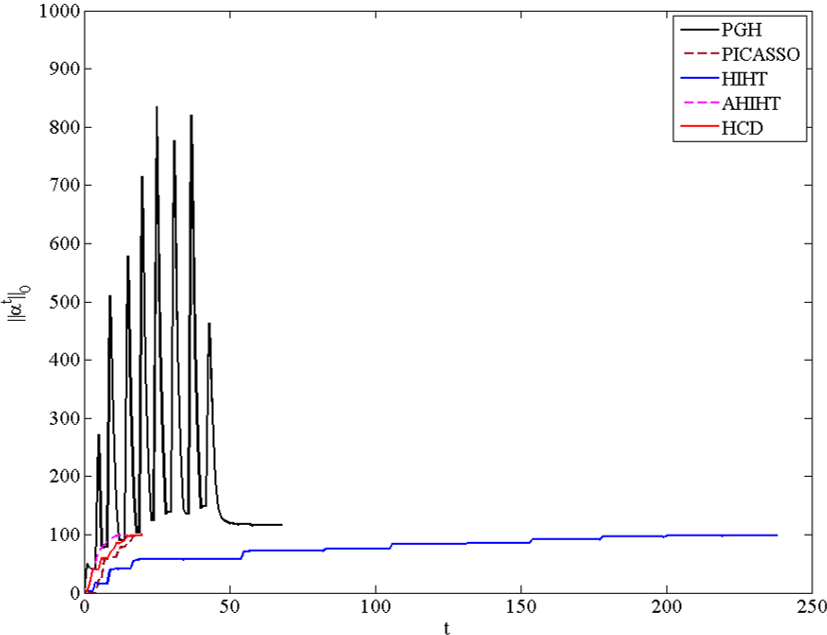}}
\subfigure[]{
\label{Fig:iteration100}\includegraphics[width=7.5cm,height=6cm]{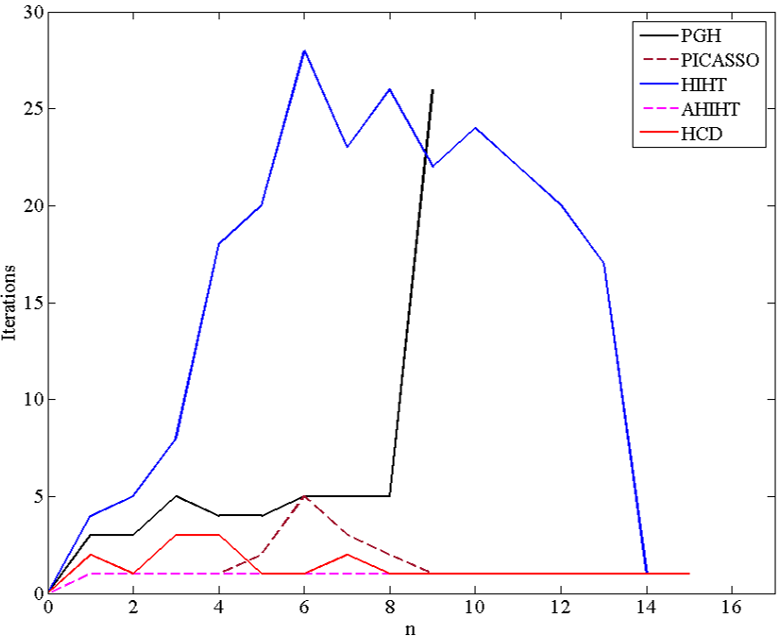}}
\caption{Performance of the compared methods on uniform distributed noisy signal. \subref{Fig:obj-gap100}: $obj\_gap$. \subref{Fig:err100}: Reconstruction error. \subref{Fig:sparsity100}: Sparsity along solution path. \subref{Fig:iteration100}: Number of iterations of each $\lambda_n$.}
\label{fig:Performance of uniform distributed noisy signal}
\end{figure*}

In this part, we show the superiority of our algorithm comparing with state-of-the-art SC algorithms.
(1) Generated signal: we generated normal distributed noisy signal with $(d=256, K=1024, s=32, \sigma=0.01)$ and uniform distributed noisy signal with $(d=1000, K=5000, s=100, \sigma=0.01)$ for comparison. In the first case, the $\lambda_{tgt}$ of HCD, HIHT and AHIHT are set as $0.01$ while it is set as $0.1$ for PICASSO and PGH in order to get sparse solution. In the second case, $\lambda_{tgt}$ is set as $0.01$ for HCD, HIHT and AHIHT, and $0.5$ for PICASSO and PGH. The performance results of each algorithm are shown in Fig.~\ref{fig:Performance of normal distributed noisy signal} and Fig.~\ref{fig:Performance of uniform distributed noisy signal}, respectively.

Fig.~\ref{Fig:obj-gap32} shows the objective gap versus the number of iterations $t$, from it we can see that HCD can obtain the lowest $ogj\_gap$ than the other four algorithms. From Fig.~\ref{Fig:err32} it can be seen that PGH obtained a much larger reconstruction error than the other four algorithms, while the other four algorithms get similar results. It can be seen from Fig.~\ref{Fig:sparsity32} that, the sparsity of the sequences $\{\boldsymbol{\alpha}^t\}$ generated by PGH and HIHT algorithms oscillate much during the iteration process. However, the sparsity generated by HCD, AHIHT and PICASSO do not oscillate, and is almost increasing with the number of iterations, they are always searcher the solution in a sparse path. Fig.~\ref{Fig:iteration32} shows the number of iterations of each $\lambda_n$. We can see that all stages of HCD and AHIHT took only 1 inner iterations and PICASSO took only 1 to 3 inner iterations to reach the relative precision, while PGH and HIHT took much inner iterations at each stage. We can make the conclusion that the two kinds of coordinate descent methods and AHIHT are more effective and efficient in reconstructing sparse representation than PGH and HIHT, while HCD and AHIHT are even better than PICASSO, this proves that $l_0$-norm is more effective than $l_p$-norm ($0<p \le 1$) in reconstructing sparse representation. Fig.~\ref{fig:Performance of uniform distributed noisy signal} demonstrates the same conclusion as Fig.~\ref{fig:Performance of normal distributed noisy signal}

(2) Natural signal: In this experiment, $\lambda_{tgt}$ is set as $0.01$ for HCD and HIHT, while for PGH and PICASSO it is set as $0.1$, and it is tuned for AHIHT to get a reasonable result. Since $\boldsymbol{\alpha}^*$ is unknown, we compare these algorithms in terms of reconstruction error, sparsity and reconstruction time, the results are shown in Tab.~\ref{tab:addlabel}. It can be seen from this table that, AHIHT fails to produce sparse representation for nature signal when the atoms of dictionary are normalized ($||\boldsymbol{d}_i||_2=1, \forall i$), while this condition is always made in practical applications to avoid trivial solution. Compared with PGH, the other three algorithms can get a more sparse solution while maintain a lower reconstruction error, indicating that the other three algorithms can obtain a better local optimal solution. Compared with HIHT, PICASSO and HCD have improved the performance of sparse coding when applied in image reconstruction, while HCD is ever better than PICASSO. In term of reconstruction time, HIHT gets the lowest computational time (expect AHIHT) due to HIHT updates all coordinates in parallel, while the computational time of our algorithm is also acceptable. For high-dimensional data, HCD has achieved a significant reduction in computational time compared with PICASSO, indicating that HCD is more suitable than PICASSO for learning the sparse representation for high-dimensional signals and more applicable in practice. Therefore, compared with the other four algorithms, our algorithm achieves a better balance between reconstruction performance and computational time, and can learn the sparse representation for natural signal more effectively.
\begin{table*}[htbp]
  \centering
  \caption{Average Results of each algorithm in natural signals}
    \begin{tabular}{c|c|c|c|c|c|c}
    \hline
    \multirow{2}{*}{Algorithm} & \multicolumn{3}{c}{Barbara \ $(d=64, K=256)$} & \multicolumn{3}{|c}{Lena \ $(d=256, K=1024)$} \\
    \cline{2-7}
    {}&{$\varepsilon $} & {$nnz$}& {Times} & {$\varepsilon $} & {$nnz$} & {Times} \\\hline
    PGH  & 0.0925 & 51.75 & 0.6463 & 0.0667 & 209.14 & 1.2278 \\\hline
    PICASSO  & 0.0238 & 44.0    & 0.0962 & 0.0329 & 146.36 & 2.5695 \\\hline
    HIHT  & 0.0428 & 46.6  & 0.0993 & 0.0388 & 161.93 & 0.3326 \\\hline
    AHIHT  & 0.8177 & 74.43  & 0.0009 & 0.1360 & 248.57 & 0.0048 \\\hline
    HCD & 0.0182 & 47.1  & 0.1074 & 0.0343 & 143.21 & 0.7613 \\\hline
    \end{tabular}%
  \label{tab:addlabel}%
\end{table*}%

\section{Conclusions}\label{Sec:Con}
This paper proposed a homotopy coordinate descent algorithm to solve the $l_0$-norm regularized least square problem in sparse coding. Differs from the classical coordinate descent algorithms, the proposed algorithm provides three strategies to speed up the convergence: warm start initialization, active set updating, and strong rule for active set initialization. Extensive computational experiments in generated signals and natural signals demonstrate that the proposed algorithm can efficiently and effectively solve the $l_0$-LS problem no matter whether the observation is noisy or not. Moreover, our algorithm perform better than four state-of-the-art homotopy methods PGH, HIHT, AHIHT and PICASSO, in both computational time and solution quality.

\bibliographystyle{IEEEtran}
\bibliography{references}
\end{document}